\documentclass[conference]{IEEEtran}
\usepackage{cite}
\usepackage{amsmath}
\usepackage{graphicx}
\usepackage{booktabs}
\usepackage{hyperref}

\begin{document}

\title{
    Train Ego-Path Detection on Railway Tracks \\
    Using End-to-End Deep Learning
}

\author{
    \IEEEauthorblockN{Thomas Laurent}
    \IEEEauthorblockA{
        \textit{FCS Railenium, F-59300 Famars, France} \\
        thomasxlaurent@gmail.com
    }
}

\maketitle

\begin{abstract}

This paper introduces the task of ``train ego-path detection'', a refined approach to railway track detection designed for intelligent onboard vision systems. Whereas existing research lacks precision and often considers all tracks within the visual field uniformly, our proposed task specifically aims to identify the train's immediate path, or ``ego-path'', within potentially complex and dynamic railway environments. Building on this, we extend the RailSem19 dataset with ego-path annotations, facilitating further research in this direction. At the heart of our study lies TEP-Net, an end-to-end deep learning framework tailored for ego-path detection, featuring a configurable model architecture, a dynamic data augmentation strategy, and a domain-specific loss function. Leveraging a regression-based approach, TEP-Net outperforms SOTA: while addressing the track detection problem in a more nuanced way than previously, our model achieves 97.5\% IoU on the test set and is faster than all existing methods. Further comparative analysis highlights the relevance of the conceptual choices behind TEP-Net, demonstrating its inherent propensity for robustness across diverse environmental conditions and operational dynamics. This work opens promising avenues for the development of intelligent driver assistance systems and autonomous train operations, paving the way toward safer and more efficient railway transportation.

\end{abstract}

The code and dataset are available at \texttt{\url{https://github.com/irtrailenium/train-ego-path-detection}}.

\section{Introduction}

In the diverse landscape of modern mobility, different transportation modes navigate their environments with varying degrees of predictability and interaction. Underground metros, for instance, operate within the confines of a closed circuit, largely isolated from unanticipated external events. Similarly, aircraft mostly adhere to predetermined flight paths with minimal unforeseen diversions. Conversely, road-based vehicles find themselves in a complex dance within a constantly evolving multi-agent environment, demanding real-time adaptability and intricate decision-making. Meanwhile, surface rail vehicles operate on an interesting middle ground. Guided by a prearranged route, dictated by the rigidity of the rails, these vehicles are bound to follow a prescribed path. However, unlike their underground metro counterpart, which are shielded from the surface world’s unpredictability, they must continuously monitor their environment. This vigilance is especially crucial for urban trams and low-speed trains, which often navigate in dynamic and densely populated areas, making their operation a blend of structured path-following and adaptive responsiveness.

Over the past few years, there has been a rapid increase in the development and deployment of intelligent systems for road vehicles. Foremost among these are the aspirations for autonomous cars and the incorporation of advanced driver-assistance systems, which have proven great effectiveness in improving road safety and driving efficiency. Building upon this precedent, there is a compelling case to be made for the integration of similar intelligent systems into the railway domain, tailoring them to the unique challenges and requirements of rail mobility.

The detection of railway tracks has emerged as a foundational task in this context, underscored by the growth in dedicated research. Most commonly, these detection methodologies rely on images captured from forward-facing cameras mounted on the train. The precise identification of railway tracks seems paramount, as it delineates the pathway the train is destined to follow, which then becomes the Region Of Interest (ROI) for subsequent systems capable of detecting and analyzing potential obstructions, anomalies or vulnerabilities.

\begin{figure}
    \centering
    \includegraphics[width=0.9\linewidth]{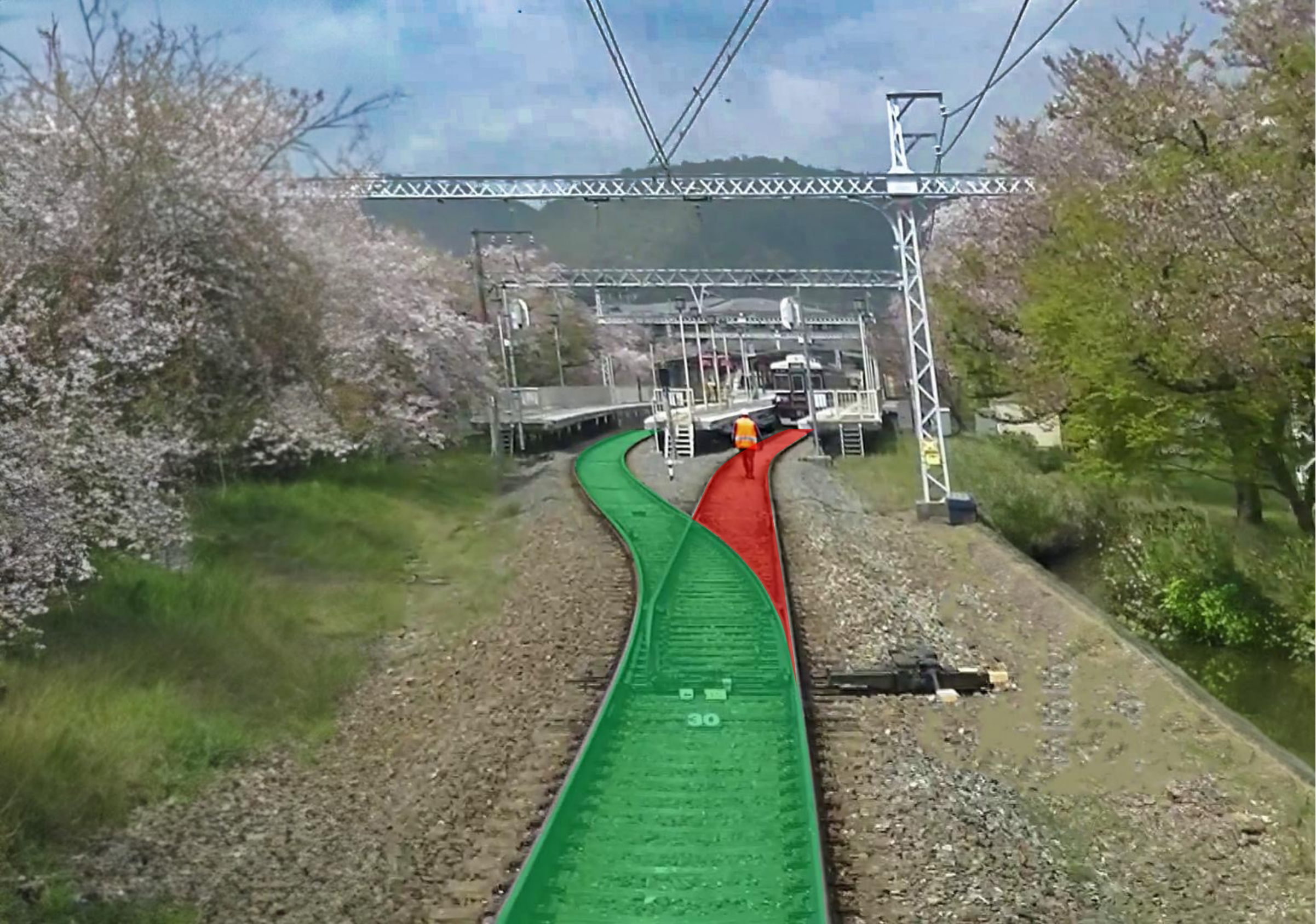}
    \caption{Ego-perspective of a train approaching a switch, thus facing a diverging track. The train’s ego-path, accurately identified using the proposed method, is highlighted in green. The track portion in red, manually annotated for illustrative purposes, is unsafe for travel in this scenario.}
    \label{fig:egopath}
\end{figure}

In modern railway infrastructures, complexity is a given: multi-track railways are commonplace, with tracks often intersecting or being guided to merge or diverge via switches. In the face of such complexities, an advanced track detection system must do more than simply detect all the rails in an image; it should identify the specific ROI that represents the train's immediate path, as illustrated in Fig. \ref{fig:egopath}. This emphasizes the significance of detecting the ``ego-path'', which identifies the train's relevant pathway amidst the potential multitude of surrounding tracks. Furthermore, given the real-time demands and integration constraints of railway operations, the ideal system would not only need to be accurate but also fast and lightweight. Equally critical is the system's robustness, which ensures seamless operation across various environmental conditions and resilience to anomalies or unexpected perturbations, like obstacles on the railway track.

In this work, we address the challenge of detecting the train's ego-path by leveraging the strengths of end-to-end deep learning methodologies. Our main contributions can be summarized as follows:

\begin{enumerate}
    \item The Train Ego-Path Detection Task: We introduce the novel concept of train ego-path detection, aimed at precisely determining the railway track the train is set to follow. We show that this refinement is crucial for enabling subsequent systems to consider the most relevant ROI, thereby reducing the risk of false positives associated with less discriminative approaches. We provide open-access ego-path annotations as an extension of the RailSem19 dataset.
    \item TEP-Net, A Tailored Deep Learning Framework: We introduce TEP-Net, a comprehensive deep learning framework that combines a configurable model architecture, a dynamic data augmentation strategy, and a domain-specific loss function to address the unique challenges of train ego-path detection. TEP-Net offers an end-to-end, decoder-free, regression-based approach that achieves remarkable accuracy and minimal latency in diverse railway environments.
    \item SOTA Paradigms Comparison: We compare TEP-Net's regression-based approach with the classification and segmentation paradigms found in state-of-the-art methods, highlighting each strategy strengths and limitations. We explain how the regression paradigm provides a superior balance between accuracy and latency, and why it allows for robust performance in dynamic and uncertain environments.
\end{enumerate}

\section{Related Work} \label{sec:related_work}

Initially, the problem of railway track detection was addressed using a multitude of conventional computer vision methods. In a pioneering work \cite{rw01}, the authors employ a dynamic programming-based method to determine the optimal pixel path in gradient images, ensuring simultaneous extraction of both left and right rails. Another notable technique is introduced by \cite{rw02}, where Inverse Perspective Mapping (IPM) is applied to railway images, allowing rails to be represented as pairs of parallel segments. Leveraging the IPM-transformed images, the authors utilize the Canny edge detector \cite{aw01} in conjunction with the Hough transform to identify rail line pairs within iterative ROIs which are then combined to generate the final rail pair. Authors of \cite{rw03} also leverage the Canny detector but opt against IPM, instead viewing rails as collections of parabolic segments from a standard perspective. These segments are recursively built from the image base to the horizon, with each new segment being chosen from a range of parabolic variations based on its predecessor. Shifting the focus toward environmental resilience, \cite{rw04} harnesses the power of HOG \cite{aw02} features for rail detection, enhancing robustness against varying weather conditions and illumination. By amalgamating similar HOG blocks using a region-growing approach, rails are delineated from the image bottom to its top. Rounding out the classic methods, \cite{rw05} employs multi-scale and multi-directional Gabor wavelet filters tailored for rail shape extraction. This method, followed by an extensive post-processing phase on the filtered image, yields the final rail detection result.

Instead of relying on predefined filters and heuristics, deep learning approaches, particularly convolutional neural networks (CNNs), have revolutionized the world of computer vision by learning optimal feature representations directly from data. The seminal work on AlexNet \cite{aw03} demonstrated the superiority of deep CNNs in image classification tasks, marking a significant shift within the scientific community toward data-driven feature extraction over traditional methods, including in the field of railway research.

\begin{figure}
    \centering
    \includegraphics[width=0.9\linewidth]{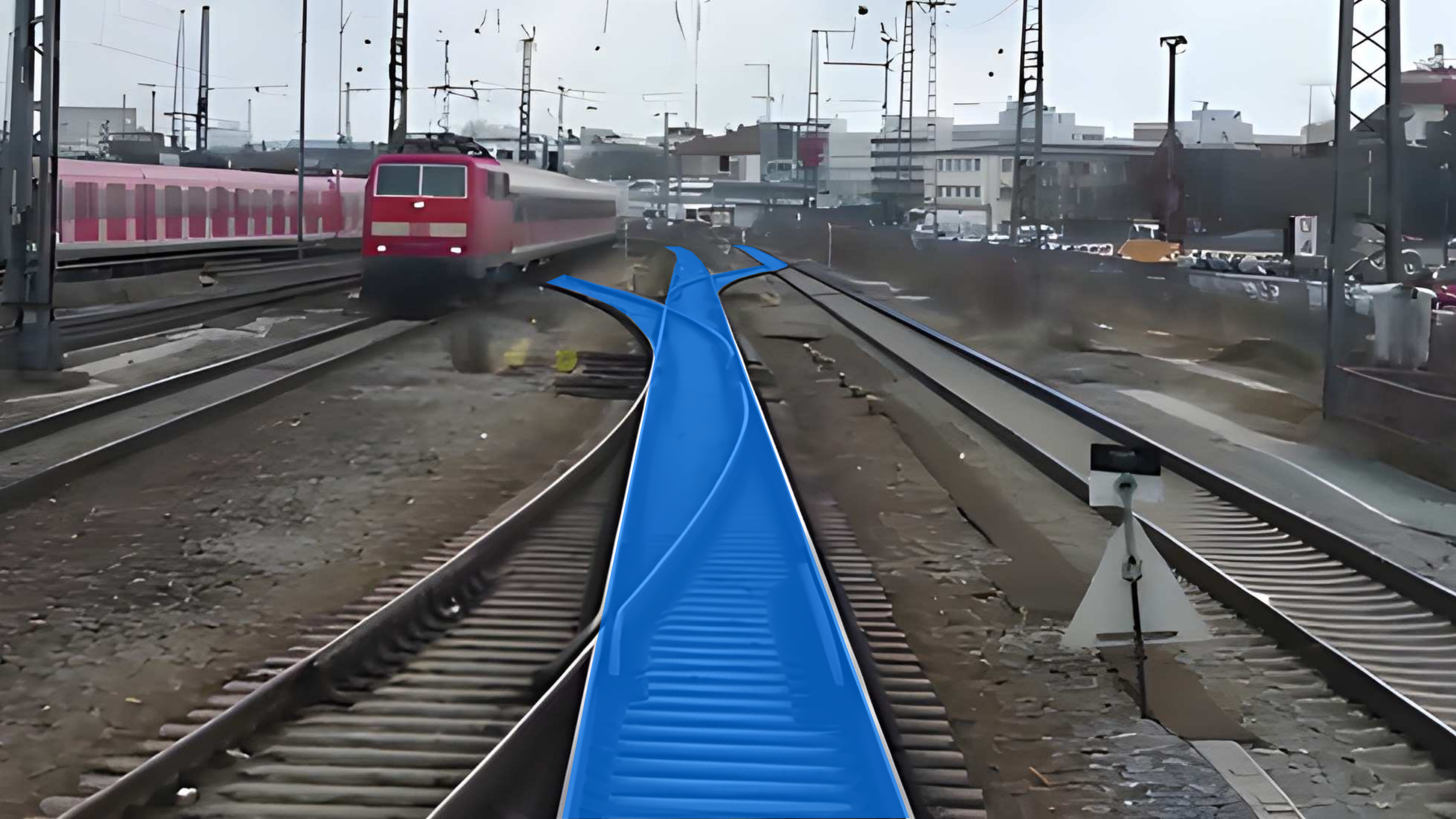}
    \caption{In this image, possible tracks are highlighted in blue. This detection area excludes non-forward-connected side tracks, thereby focusing on a more relevant ROI.}
    \label{fig:possible_tracks}
\end{figure}

The paradigm shift began with \cite{rw06}, in which authors leverage a custom CNN inspired by SegNet \cite{aw04}, incorporating atrous spatial pyramid pooling \cite{aw05} along with mixed pooling \cite{aw06}. The challenge of identifying railway tracks within the image is addressed through an objective of semantic segmentation, i.e., classifying each pixel as either track or no track. The segmentation outputs are further refined with a polygon fitting method to produce the final smoothed railway track area. Following suit, \cite{rw07} presents RailNet, a segmentation model based on the fully convolutional networks architecture \cite{aw07}, with the ResNet-50 \cite{aw08} backbone. This approach leverages multi-scale feature aggregation via pyramidal feature propagation \cite{aw09} of summed feature maps from each ResNet stage, leading to superior segmentation capabilities. Further contributions in \cite{rw08} do not significantly deviate in terms of the segmentation network, utilizing a custom U-Net-based architecture \cite{aw10} that incorporates the ResNet-34 backbone with a spatial pyramid pooling module \cite{aw11} and a squeeze-and-excitation self-attention mechanism \cite{aw12} on the skip connections. However, the novelty lies in the concept of ``possible tracks'' extraction: assuming that switch states cannot be precisely determined, possible tracks are the tracks that are forward-connected to the train's running track, as illustrated in Fig. \ref{fig:possible_tracks}. The authors propose a rule-based algorithm that effectively extracts possible tracks from the segmented image by extrapolating, in a bottom-up process, the most centered track borders and by identifying and processing track intersections. In the same quest to detect possible tracks, \cite{rw09} proposes an approach utilizing a segmentation network inspired from fully convolutional DenseNets \cite{aw13}. This method goes beyond standard segmentation by training the network to produce a regression output, i.e., a 1-channel image where pixel intensity is related to the probability of the corresponding point to be in the center of a track. This quite intricate approach, supplemented by point clustering and a substantial post-processing treatment, allows for a comprehensive, however complex strategy for possible tracks extraction. Distinctively, \cite{rw10} departs from the segmentation paradigm by converting the task of track detection into a series of row-wise multi-class classification problems. Specifically, the image is divided into grids (one for each rail), and for each row of the grids, the target class is the cell containing the target rail. This method, initially proposed for road lane line detection \cite{sw01}, allows bypassing the need for a decoder in the network, instead relying on the global features extracted by the encoder.

\section{Train Ego-Path Detection}

\subsection{Motivation}

Traditional computer vision detection methods, on the one hand, are conceptually constrained to detect a single track instance on an image, relying on hard-coded knowledge of the track's geometrical and topological characteristics. However, these methods often lack environmental robustness and struggle to accommodate complex railway infrastructures. On the other hand, deep learning-based segmentation methods offer a more flexible and robust solution for railway track detection by not being restricted by specific constraints. These methods, however, detect railway tracks across the entire image without any distinction between instances, which may not be desirable. Although some deep learning methods have attempted to isolate possible tracks, this refinement of the ROI, although significant, is not sufficiently discerning. This is demonstrated in Fig. \ref{fig:egopath} and Fig. \ref{fig:possible_tracks} where a possible track could lead directly into an obstruction. The ambiguity resulting from this imprecision could result in a significant increase in the number of false positives in subsequent systems (e.g., obstacle detection). These might misinterpret non-threatening elements, due to the lack of distinction between possible tracks and the actual path the train is anticipated to follow. 

Assuming that the real-time visual data provides sufficient information to determine the precise railway track on which the train is set to run, we introduce the concept of ``train ego-path detection''. Therefore, the primary objective of this task is to encourage the development of methods that can detect the most relevant ROI for subsequent systems to make better-informed decisions.

\subsection{Dataset}

For our study, we utilize the RailSem19 dataset \cite{data}, which contains 8500 images with a wide range of annotations. This dataset covers a broad spectrum of conditions, featuring images from 530 distinct video sequences and covering more than 350 hours of train and tram traffic across 38 countries. It captures all four seasons along with diverse weather and lighting conditions, and includes footage from a range of camera setups with different fields of view, mounted in various positions on the front of the train. RailSem19 is the most popular dataset in the field of computer vision for railway intelligent systems and is particularly suitable for training a robust and universal model for ego-path detection.

Among the annotations of Railem19 are the rail annotations, which are provided as polylines, i.e., as a series of 2D points within the image space. These polylines are grouped in pairs to represent individual track instances, with a left and a right rail. However, a key limitation of the original dataset is the lack of specific identification of the ego-path, i.e., the pair of rails on which the train is set to travel. To meet our specific needs, we extend the dataset by adding an annotation that explicitly identifies the train's ego-path. For most images in the dataset, totaling 7917, we introduce a polyline pair that specifies the left and right rails of the ego-path. The remaining 583 images are not annotated because they either depict a very unusual point of view or present situations where the ego-path cannot be determined with certainty (refer to subsection \ref{sub:limitations} for more details).

\begin{figure}
    \centering
    \includegraphics[width=0.75\linewidth]{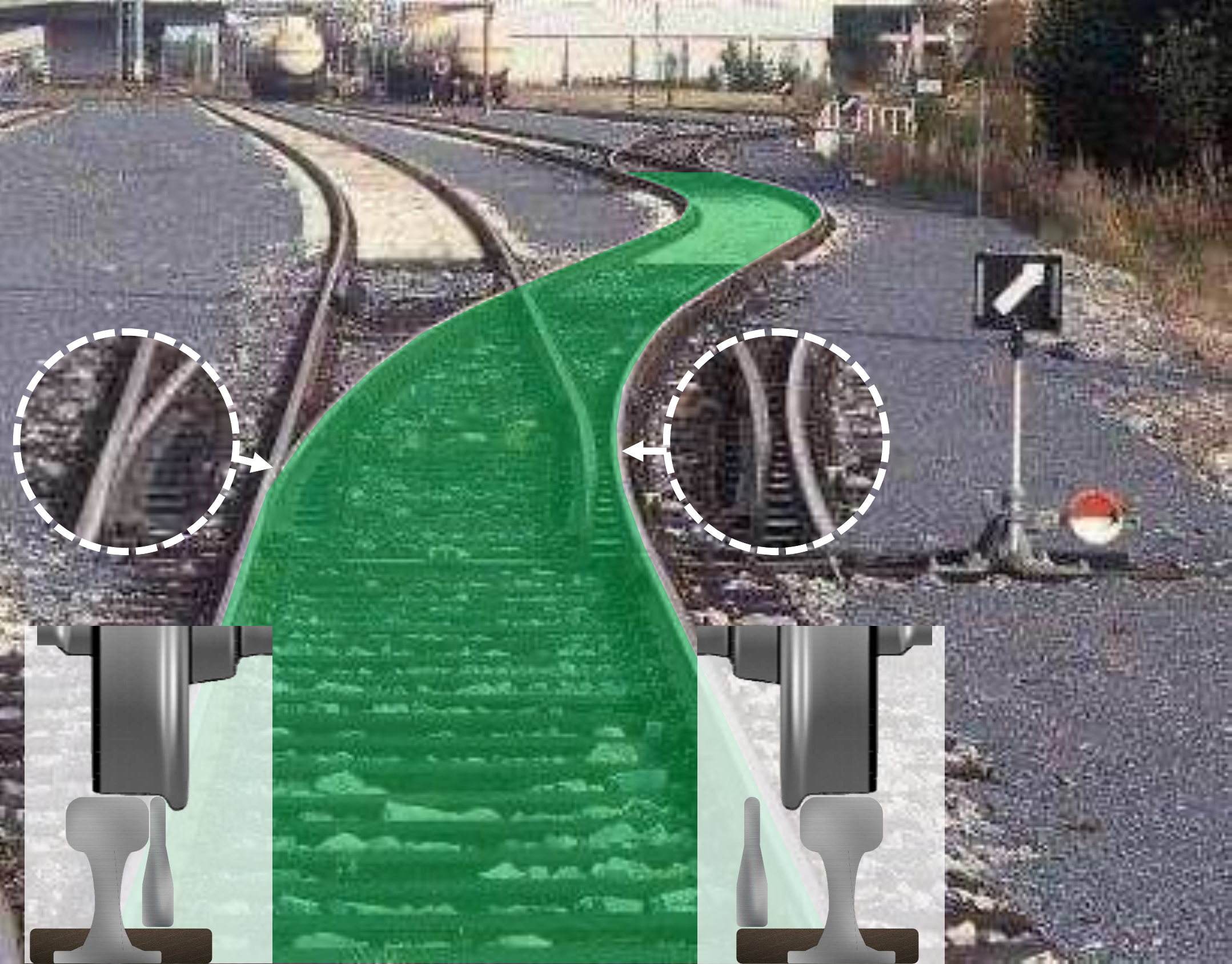}
    \caption{Close-up view of a railway switch mechanism. The switch blades configuration is highlighted: closed on the left side and open on the right side. The bottom wheel-rail schematics illustrate the train's wheels, featuring a special profile with a flange, interacting with the switch blades. This clarifies why a train approaching this switch will be routed to the right, as confirmed by the directional arrow on the sign standing on the right of the image.}
    \label{fig:switch}
\end{figure}

The annotation process is largely facilitated by the existing rail annotations in RailSem19. For most images, the automated selection of the rail pair most aligned with the camera's optical axis adequately identifies the ego-path. However, some cases require human intervention due to issues such as incorrect rail pairing, discontinuities in rail annotations, or inconsistencies when tracks intersect. In these cases, manual annotation is necessary, performed using the CVAT annotation tool. When dealing with images that present railway switches, annotation requires an understanding of the mechanism to be able to deduct the ego-path from the switch state, as explained in Fig. \ref{fig:switch}. Besides that, our guiding principle is to annotate only the ego-path portions that are definitively known. For instance, if the state of a switch is unclear, as shown at the top of Fig. \ref{fig:switch}, the annotated path terminates at that point.

\section{Methodology}

\subsection{Data Augmentation} \label{sub:data_augmentation}

\begin{figure}
    \centering
    \includegraphics[width=0.9\linewidth]{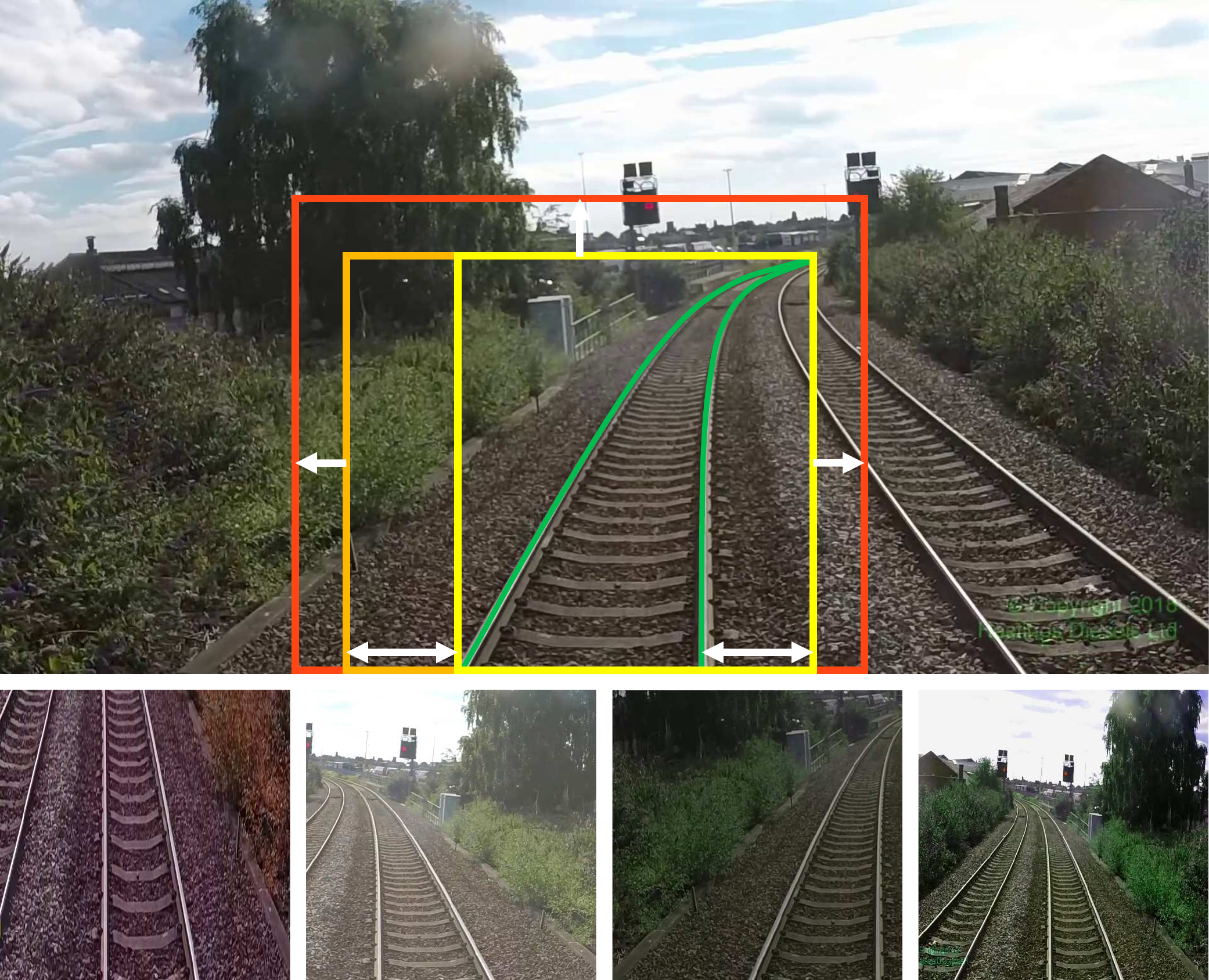}
    \caption{Illustration of our data augmentation strategy. Green polylines delineate the ego-path ground truth, encapsulated by the yellow rectangle that defines the base ROI. The orange extension expands the ROI to place the base of the rails at its center, while the red rectangle adds predefined margins. To this resulting ROI, variability is introduced by applying stochastic shifts to the left, top, and right borders, conforming to a normal distribution with tunable parameters. This introduced variability is subject to constraints ensuring that the rail base remains entirely within the resulting images. Four examples of augmented versions of the top image are shown in the bottom row.}
    \label{fig:data_augmentation}
\end{figure}

Within our ego-path detection framework, we introduce a data augmentation strategy tailored to the commonly large field of view of standard train-mounted camera setups, which often captures substantial areas irrelevant to track detection. To optimize the model's efficiency and precision, our augmentation process strategically targets the image region where the rails are predominantly visible. This region must account for variability since the track location can significantly vary within the frame, such as a leftward shift in images of left turns, as shown in Fig. \ref{fig:bounded_roi}. To address this variability and accurately mirror real-life conditions, our augmentation policy not only crops the image to the ego-path ROI but also introduces random margins. This methodology, explained in Fig. \ref{fig:data_augmentation}, also includes conventional image adjustments (i.e, brightness, contrast, saturation, and hue variations) and random horizontal flips. As a result, the same image leads to highly diversified outcomes when augmented. This strategy is applied in an online fashion: at each epoch of the training process, the augmented version of an image is randomly generated.

\subsection{Problem Formulation}

As detailed in section \ref{sec:related_work}, current deep learning approaches for railway track detection adopt segmentation-based methods to detect multiple tracks in the input image. In this work, we shift our focus to ego-path detection, which constrains the detection to a single track per image. In this context, we propose a novel formulation grounded in the regression paradigm, which aligns more naturally with the problem at hand. Regression-based methods have shown promising results in analogous applications within autonomous driving, such as autonomous car trajectory planning \cite{sw02} and road lane line detection \cite{sw03}, showcasing accuracy, robustness, and computational efficiency.

\begin{figure*}
    \centering
    \includegraphics[width=0.9\linewidth]{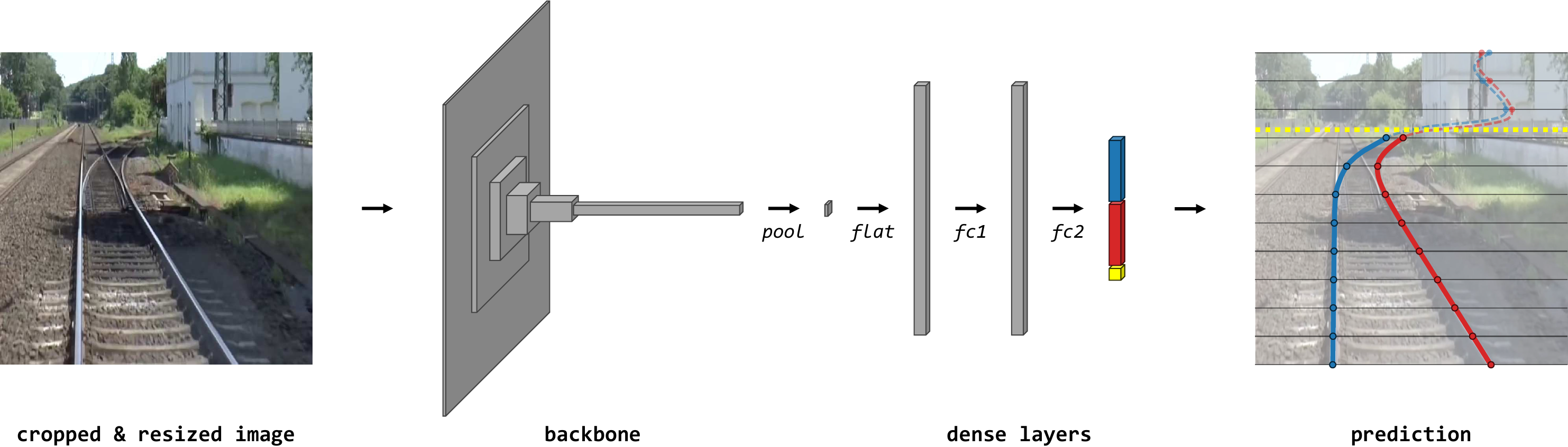}
    \caption{TEP-Net architecture and prediction methodology: starting from a cropped and resized input image, features are extracted via the backbone, passed through dense layers for regressing the output values, which are then processed to produce the final ego-path prediction.}
    \label{fig:tepnet}
\end{figure*}

Specifically, we define the regression problem as follows: given an image, the objective is to regress the position of the ego-path rail pair within the image space. This approach ensures both the existence and uniqueness of the solution, which segmentation-based methods do not guarantee. Conceptually, a rail can be visualized as a curve within the image plane. One approach to modeling such a curve could involve hypothesizing that rails correspond to second or third-degree polynomial curves and regressing the coefficients of these equations. However, rails can display a variety of geometric configurations, some of which cannot be accurately fitted by polynomials. While more sophisticated methodologies, such as spline interpolation, might offer closer approximations, they introduce complexity into the modeling process. To avoid making any presumptions about the rail's shape, our proposed method instead regresses a set of points in the image space which, when connected, form a curve defining the rail. As illustrated in the prediction overlay of Fig. \ref{fig:tepnet}, this approach forms the core of our methodology, wherein we regress the $x$-axis values for a series of $y$-axis values, or ``anchors'', along both the left and right rails. Given that rail curves do not necessarily intersect the cropped images' top edge, it is also essential to regress the ego-path $y$-limit, i.e., the highest $y$-axis value of the ego-path in the image space. Rail points aligned with anchors above this threshold are not included in the final ego-path predictions, as they do not contribute relevant information for most applications.

\subsection{TEP-Net Model} \label{sub:tep_net}

With TEP-Net (Train Ego-Path Network), we aim to offer a general approach for employing deep CNNs for ego-path detection in a regression fashion, rather than prescribing a rigid architecture. Unlike semantic segmentation networks, which typically feature an encoder-decoder structure, the TEP-Net model employs a conventional architecture consisting of an encoder followed by a single head of fully connected layers that directly regresses the outputs, as illustrated in Fig. \ref{fig:tepnet}.

The encoder can be chosen from established CNN backbones such as ResNet, EfficientNet \cite{aw14}, or others, in accordance with specific application requirements and constraints. Its primary function is to extract essential features from the input image, yielding an output with reduced spatial dimensions but a significantly increased number of channels. These feature maps are then processed through a 1\texttimes1 convolutional pooling layer, that substantially reduces the channel count, yielding a manageable feature set for the subsequent layers. Indeed, the pooled feature maps are then flattened and sequentially passed through a concise series of fully connected layers, typically two, resulting in a prediction vector of size $2 \times H + 1$, where $H$ denotes the number of anchors. This vector contains all the necessary regressed values for the ego-path prediction: $H$ for the left rail's $x$-values, another $H$ for the right rail's $x$-values, and a last value for the ego-path $y$-limit.

Our design ensures that the $y$-limit value is bounded between 0 and 1, signifying the image's vertical boundaries, through a Sigmoid function. However, the $x$-values are left unconstrained, allowing predictions beyond the image's horizontal scope. This unbounded nature is indeed crucial for the network to interpret and understand the trajectory of rails in the real world, which sometimes extends beyond the limited perspective of a single cropped image. Our experiments confirm that this freedom significantly enhances the model's precision.

TEP-Net is the outcome of extensive experimentation, which culminates in a remarkably simple yet highly effective architecture for ego-path detection. Moreover, TEP-Net is an end-to-end framework, meaning that a single neural network is responsible for processing the raw input data through to the final prediction vector. This conceptual simplicity facilitates implementation and customization by eliminating the need for complex post-processing or intermediary steps, enabling direct training and inference within a unified model structure.

\subsection{Loss Function}

In the pursuit of optimizing the TEP-Net model, we define a tailored loss function, grounded in the regression paradigm, that aligns with our problem formulation. This function further integrates real-world spatial considerations, in order to produce models that are both accurate and consistent for application in ego-path detection. 

Let matrices $\widehat{\mathbf{X}}$ and $\mathbf{X}$ be the predicted and actual $x$-values at the intersections of the anchors with the left and right rails, and let $\widehat{y_{lim}}$ and $y_{lim}$ be the predicted and actual values of the ego-path $y$-limit. All values are relative to image dimensions.

\subsubsection{Trajectory Loss}

The first component of our loss function is the trajectory loss, $L_{traj}$, which measures the lateral deviation between the predicted and actual ego-rails:

\begin{equation} \label{eq:ltraj}
    L_{traj} = \frac{\sum_{i=1}^{H} m_i \times w_i \times L_{rails}(\widehat{\mathbf{X}}_i, \mathbf{X}_i)}{\sum_{i=1}^{H} m_i}
\end{equation}

\noindent in which

\begin{equation} \tag{1.1}
    L_{rails}(\widehat{\mathbf{X}}_i, \mathbf{X}_i) = \sum_{r \in \{\text{left}, \text{right}\}} \mathit{SmoothL1}(\widehat{\mathbf{X}}_{i,r} - \mathbf{X}_{i,r}, \beta_{1})
\end{equation}

with

\begin{equation} \tag{1.1.1}
    \mathit{SmoothL1}(x, \beta) =
    \begin{cases}
        0.5x^2 / \beta & \text{if } |x| < \beta \\
        |x| - 0.5 * \beta & \text{otherwise},
    \end{cases}
\end{equation}

\begin{equation} \tag{1.2}
    w_i = \min\left(\frac{1}{\mathbf{X}_{i,right} - \mathbf{X}_{i,left}}, W_{max}\right),
\end{equation}

\begin{equation} \tag{1.3}
    m_i =
    \begin{cases}
        1 & \text{if } i \leq y_{lim} \times H \\
        0 & \text{otherwise}
    \end{cases}
\end{equation}

As formulated in the numerator of \eqref{eq:ltraj}, the trajectory loss sums, along the image anchors, the smoothed L1 errors between the predicted and actual $x$-values of the ego-rails points. We use the smooth L1 criterion to ensure that our loss is linearly proportional to the magnitude of the error, incorporating a smooth transition around zero to accommodate the inherent noise and imprecision in data annotation. For this purpose, we fix $\beta_{1} = 0.005$.

These errors are weighted by a $w_i$ factor, inversely proportional to the rail width, in order to compensate for the linear perspective effect in images: as the rails converge toward the vanishing point of an image, the same error in pixel measurement corresponds to a linearly increasing real-world distance. We use $W_{max}$ as a constant upper bound of $w_i$ to prevent abusive weighting. In our experiments, we set $W_{max}$ as the 95\textsuperscript{th} percentile of the unbounded $w_i$ values obtained from the training set. With the data augmentation configuration employed, the resulting value for $W_{max}$ is around 20.

A masking factor $m_i$ is applied to zero out the contribution of anchors above the ego-path $y$-limit, since the target points beyond this limit are arbitrarily initialized and do not represent meaningful data. The sum of the $m_i$ values acts as a normalizing factor in the denominator of \eqref{eq:ltraj}, ensuring that the loss is averaged only over the relevant portion of the ego-path.

\subsubsection{Y-Limit Loss}

Alongside $L_{traj}$, we set $L_{ylim}$: 

\begin{equation}
    L_{ylim} = \mathit{SmoothL1}(\widehat{y_{lim}} - y_{lim}, \beta_{2})
\end{equation}

We also use the smooth L1 loss to quantify the error between the predicted and actual ego-path $y$-limit. This time, we set $\beta_{2} = 0.015$ to account for the increased imprecision in data annotation associated with image areas that are often significantly blurred, especially when the ego-path disappears into the horizon.

\subsubsection{Composite Loss Function}

The final loss function, $L$, combines the ego-path trajectory and $y$-limit losses:

\begin{equation}
    L = L_{traj} + \lambda \times L_{ylim}
\end{equation}

Here, $\lambda$ is the coefficient that adjusts the relative importance of the $y$-limit term in the overall loss calculation. Empirically, we find that a value of $\lambda = 0.5$ provides an effective balance, ensuring that both components of the ego-path prediction are efficiently optimized during the network training phase.

\section{Experiments} \label{sec:experiments}

\subsection{Experimental Setup}

\subsubsection{Model}

Based on the global TEP-Net architecture detailed in subsection \ref{sub:tep_net}, we fix a set of parameters for our experimental models, aiming for a balance between performance and computational efficiency. The input shape is 3\texttimes512\texttimes512 which, on average, is comparable to the original image resolution post-cropping. We evaluate TEP-Net across a range of widely-used backbone architectures, specifically ResNet (RN) versions 18, 34, 50, and EfficientNet (EN) versions B0, B1, B2, B3. The convolutional pooling layer employs 8 filters, yielding a feature map of 8\texttimes16\texttimes16 (2048-dimensional flattened feature vector) across all backbone variants. The fully connected part of the network includes a single hidden layer of 2048 neurons with the ReLU activation function. The output vector size is 129, due to our use of $H = 64$ horizontal anchors, which is sufficient for a smooth rail profile.

\subsubsection{Training}

For our experiments, we randomly sample 80\% of the dataset for training, 10\% for validation, and the remaining 10\% for testing. We utilize the Adam optimizer and employ PyTorch's OneCycle learning rate policy, which initially increases the learning rate up to 0.0001 and then gradually decreases it back to 0. Our model training is conducted with a batch size of 8 over 400 epochs. These hyperparameters result from a grid search aimed at minimizing the final validation loss. We do not employ early stopping, but we instead select the checkpoint associated with the lowest validation loss from the final 10\% of training epochs. For the input data, we use RGB images augmented following the strategy detailed in subsection \ref{sub:data_augmentation} and resized to 512\texttimes512 pixels. We train our models with PyTorch on a single NVIDIA RTX A6000 GPU. All training configurations, logs and results are available on \texttt{\url{https://wandb.ai/greattommy/train-ego-path-detection}}.

\subsubsection{Evaluation}

For model evaluation, we employ the same cropping procedure as used during the training phase. We believe that this approach is the fairest, as it is aligned with practical usage where the model processes cropped images (refer to subsection \ref{sub:img_cropping} for more details). The test images are not altered in any other way, meaning they do not undergo conventional image adjustments. Our assessment focuses on two critical performance indicators: accuracy and latency. Accuracy is quantitatively evaluated using the Intersection over Union (IoU) metric between predicted and actual ego-paths. For the computation of the IoU, predicted ego-paths are transformed into binary masks by connecting the points that constitute them with straight lines and filling the resulting shape. The target masks are directly derived from the data points of the annotations to ensure maximum precision; predicted ego-paths are adjusted to align with the target dimensions. Although the IoU metric not accounting for the linear perspective effect when evaluating accuracy, we choose it for its simplicity and widespread use in computer vision tasks, including railway track detection. Latency, a crucial factor for real-time applications, is measured on both PyTorch (PT) and TensorRT (TRT) runtimes. Specifically, we measure the duration of the neural network inference, deliberately excluding pre- and post-processing times, as they largely depend on their particular implementation. Regarding inference precision, we opt for full precision (fp32) with PyTorch, as it shows smaller latency than half precision (fp16). For TensorRT, we use automatic mixed precision, mixing fp32 and fp16, and effectively maintaining full accuracy, as measured experimentally. Benchmarking is performed on an NVIDIA Quadro RTX 6000 GPU, with PyTorch v2.2.1 and TensorRT v8.6.1 (CUDA v12.1).

\subsection{Quantitative Results}

\begin{table}
    \centering
    \caption{Performance of the TEP-Net Model}
    \begin{tabular}{lcccc}
        \toprule
        \textbf{Model} & \textbf{Param.} & \textbf{MACs} & \textbf{Latency PT / TRT} & \textbf{IoU}    \\
        \midrule
        TEP-ENB0       & 8.06M           & 1.91G         & 9.02 / 1.10 ms            & 0.9730          \\
        TEP-ENB1       & 10.56M          & 2.87G         & 12.89 / 1.60 ms           & 0.9734          \\
        TEP-ENB2       & 11.67M          & 3.31G         & 13.08 / 1.67 ms           & 0.9748          \\
        TEP-ENB3       & 14.57M          & 4.87G         & 14.90 / 2.08 ms           & \textbf{0.9753} \\
        TEP-RN18       & 15.64M          & 9.48G         & \textbf{2.30 / 0.60 ms}   & 0.9695          \\
        TEP-RN34       & 25.75M          & 19.14G        & 3.98 / 0.99 ms            & 0.9745          \\
        TEP-RN50       & 27.99M          & 21.36G        & 6.04 / 1.28 ms            & 0.9723          \\
        \midrule
        ResNet-50      & 25.56M          & 21.36G        & 6.12 / 1.28 ms            & -               \\
        \bottomrule
    \end{tabular}
    \label{tab:tepnet}
\end{table}

\begin{figure*}
    \centering
    \includegraphics[width=0.9\linewidth]{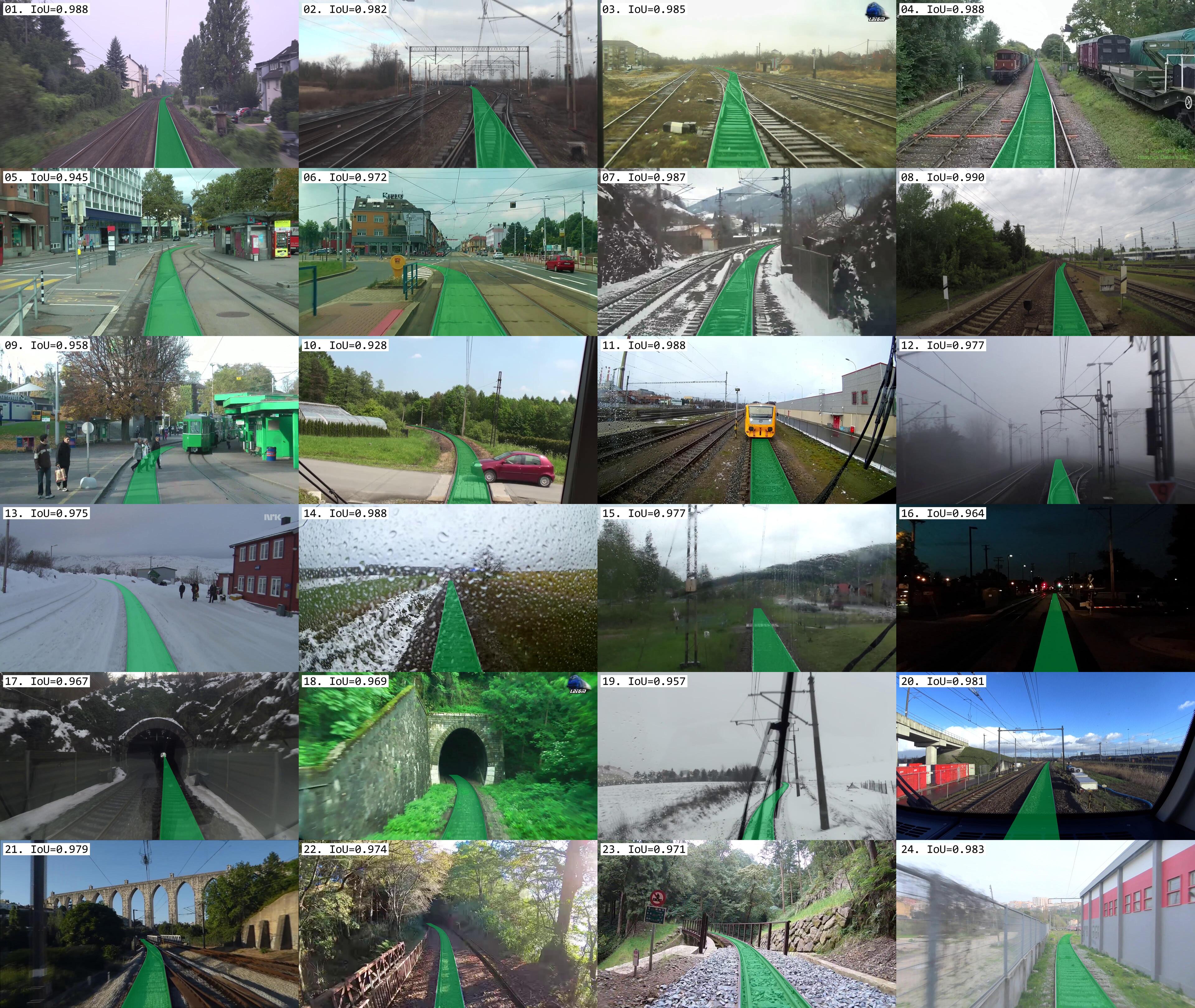}
    \caption{Qualitative results of the TEP-Net model using the EfficientNet-B3 backbone.}
    \label{fig:qualitative_tepnet}
\end{figure*}

The TEP-Net model's quantitative assessment, presented in Table \ref{tab:tepnet}, demonstrates very strong results in ego-path detection, both in terms of accuracy and latency. Performance is put into perspective with the model's parameter count (Param.) and the number of Multiply-ACcumulate (MAC) operations, which vary based on the chosen backbone. IoU scores span from 0.9695 with the ResNet-18 backbone to 0.9753 with the EfficientNet-B3 backbone. Remarkably, TEP-Net shows minimal latency, particularly when using the TensorRT runtime, achieving a speed increase of more than sixfold compared to PyTorch. With the ResNet-18 as backbone and TensorRT, TEP-Net reaches an impressive frame rate of 1667 frames per second, all while maintaining high accuracy.

Interestingly, TEP-Net versions equipped with EfficientNet backbones exhibit higher latencies compared to their ResNet counterparts, despite having fewer parameters and MACs. This can be attributed to the presence of a depth-wise convolution layer in EfficientNet's MBConv blocks, as depth-wise convolutions are inherently slower than traditional convolutions for the same parameter count.

Additionally, the analysis of an unmodified ResNet-50 model, as detailed in the bottom row of Table \ref{tab:tepnet}, reveals only a slight increase in parameters associated with the TEP-Net architecture, due to its deeper fully connected segment. However, MACs and latencies remain almost unaffected, emphasizing that the computational demand primarily lies in the unaltered backbone convolutions. In conclusion, TEP-Net's streamlined architecture ensures that its complexity is largely attributed to the backbone's feature extraction capabilities, making it challenging to envisage a simpler approach to ego-path detection without compromising effectiveness.

\begin{figure*}
    \centering
    \includegraphics[width=0.9\linewidth]{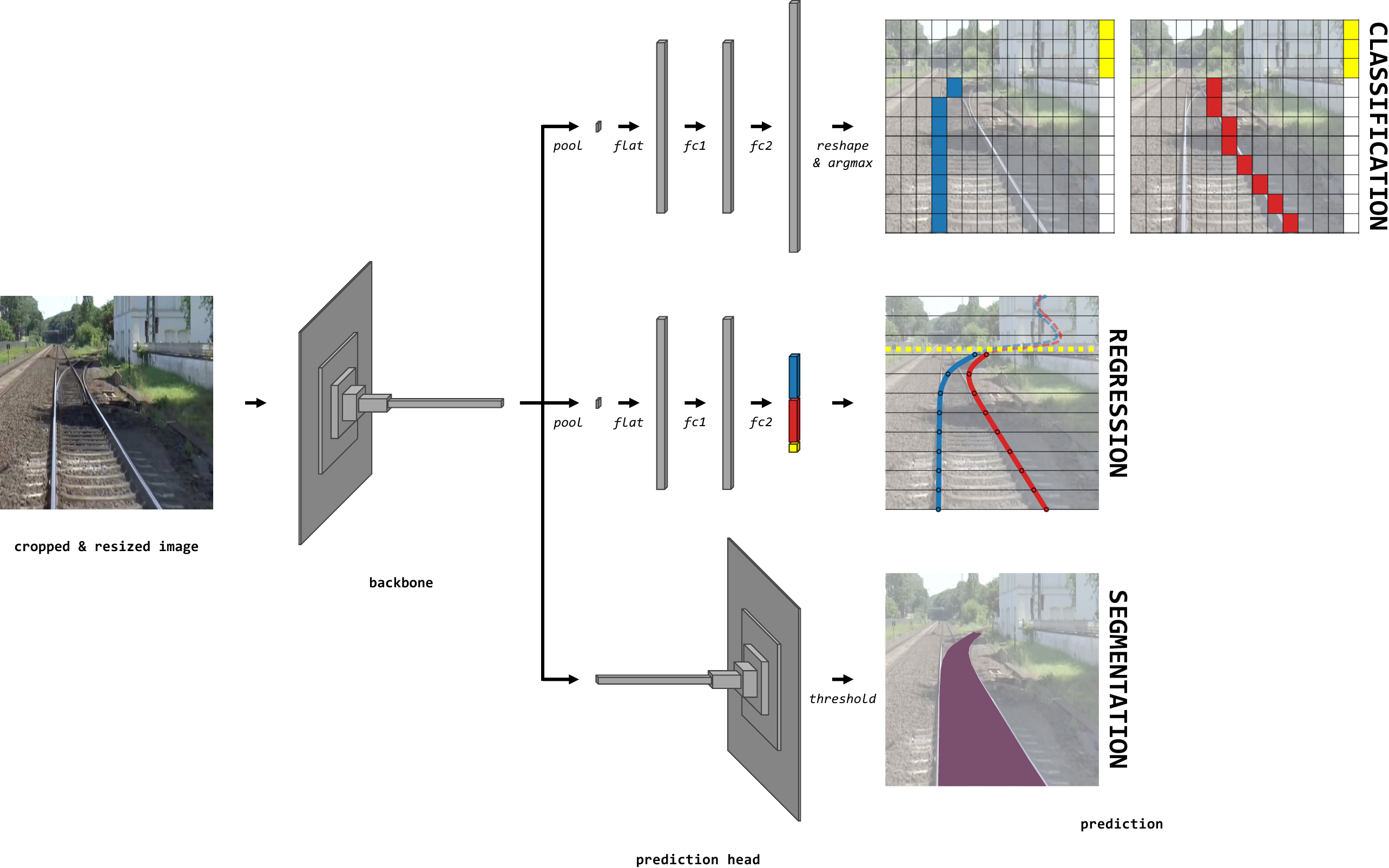}
    \caption{TEP-Net architectural variants to adapt to classification, regression and segmentation paradigms.}
    \label{fig:paradigms}
\end{figure*}

\subsection{Qualitative Results}

The qualitative assessment of TEP-Net's performance in ego-path detection across a multitude of scenarios is depicted in Fig. \ref{fig:qualitative_tepnet}. Image 1 exemplifies a straightforward rail environment where TEP-Net very precisely detects the ego-path amidst minimal track complexity and optimal visibility. Intricate multi-track configurations with intersecting rails, as observed in images 2 and 3, do not impede the model's ability to identify the ego-path, achieving an IoU exceeding 0.98 in both cases. Notably, without relying on any hard-coded prior knowledge about rail configurations, TEP-Net has adeptly learned from the data to read and interpret switch states, thereby accurately determining the correct ego-path when facing situations with merging tracks (images 4, 21), or diverging tracks (images 5-8, 12, 24). Leveraging the diversity of scenarios present in RailSem19, TEP-Net is able to detect the ego-path for various rail vehicles: it works as well for regular trains as for urban trams and their distinctive infrastructural elements (images 5, 6, 9). Remarkably, TEP-Net shows robustness against obstructions, seamlessly predicting paths despite partial occlusions by pedestrians (image 9) or vehicles (image 10), and adheres to annotation guidelines by not overpredicting when facing complete occlusion (images 11, 18). The model's resilience is further highlighted in its handling of diverse weather and lighting conditions, maintaining consistent performance in snow (images 7, 13, 19), rain (images 14, 15), and nighttime settings (images 16). Regarding tunnels, TEP-Net has the correct behavior by predicting the ego-path through the tunnel when the exit is visible (image 17) and stopping at the tunnel entrance when it is not (image 18). TEP-Net accommodates various camera placements, including windshield-mounted views, remaining unaffected by elements such as windshield wipers (image 19) or cab desks (image 20). The model also handles unusual rails conditions like corrosion (image 21), and is not confused by guardrails (images 22, 23) or any other infrastructure that presents rail-like patterns (image 24). Variations in rail width or perspective (images 22, 23) once again present no discernible challenge.

In essence, TEP-Net is able to deliver accurate ego-path detection across a wide spectrum of scenarios, affirming its suitability for practical rail navigation applications. When trained on RailSem19, the model can be used out of the box, without the need for further adjustments or additional data, and will work across a variety of environments, rail vehicles, and camera setups.

\section{SOTA Paradigms Comparison}

Building on the compelling quantitative and qualitative results previously discussed, it is clear that TEP-Net presents an efficient and robust solution for ego-path detection. But how does it compare to existing state-of-the-art (SOTA) methods? Is it the most effective solution for this task? Addressing this question is not straightforward, given that TEP-Net is trained on a novel task with its dedicated dataset, making direct comparisons less relevant. Therefore, we decided to evaluate the regression-based approach used by TEP-Net against the paradigms behind the leading SOTA track detection methods.

\subsection{Experimental Setup}

Predominantly, SOTA methods employ segmentation networks with varying complexities in their encoder-decoder architectures, occasionally incorporating post-processing stages. An exception is noted in \cite{rw10}, where the problem is addressed through row-based classification on a prediction grid that is fully connected to the encoder output features. Consequently, we decided to compare TEP-Net, which lies in the regression paradigm, against both the classification and segmentation paradigms. In order to conduct the comparison in the fairest manner possible, we implemented these paradigms by adapting the TEP-Net architecture.

Fig. \ref{fig:paradigms} illustrates how TEP-Net's prediction head can be interchanged to accommodate the classification and segmentation paradigms. For the classification paradigm, the prediction head mirrors that of the regression paradigm, except for the final fully connected layer, whose output size is significantly larger. The output vector indeed needs to be reshaped into a prediction grid with dimensions $C \times H \times (W + 1)$. Here, $C$ is the number of rails to be detected (set to 2 as we detect a unique ego-path), $H$ (set to 64, same as for regression) represents the horizontal anchor count, and $W$ (set to 128, based on the optimal performance setting from \cite{rw10}) denotes the class count, i.e., the number of columns in the prediction grids. Following the methodology in \cite{rw10}, one column is added for the background class, allowing the network to predict the absence of rails. For the segmentation paradigm, we adopt a U-Net-like architecture, chosen for its proven effectiveness in various segmentation tasks. The prediction head consists of a decoder that sequentially upsamples the feature maps and merges them with skip connections from the encoder stages, ultimately yielding a binary mask of shape 1\texttimes512\texttimes512.

The loss functions for these paradigms are tailored accordingly: the binary Dice loss is employed for segmentation, ensuring overlap maximization, while the cross-entropy loss is used for classification, optimizing class probability distributions. The backbone architectures, data augmentation strategy, training procedures and hyperparameters remain consistent across models, except for paradigm-specific parameters and for the epoch count, which is set to 300 for segmentation and to 200 for classification, accounting for their faster convergence rates in comparison to the regression approach.

\subsection{Results}

\begin{figure}
    \centering
    \includegraphics[width=0.9\linewidth]{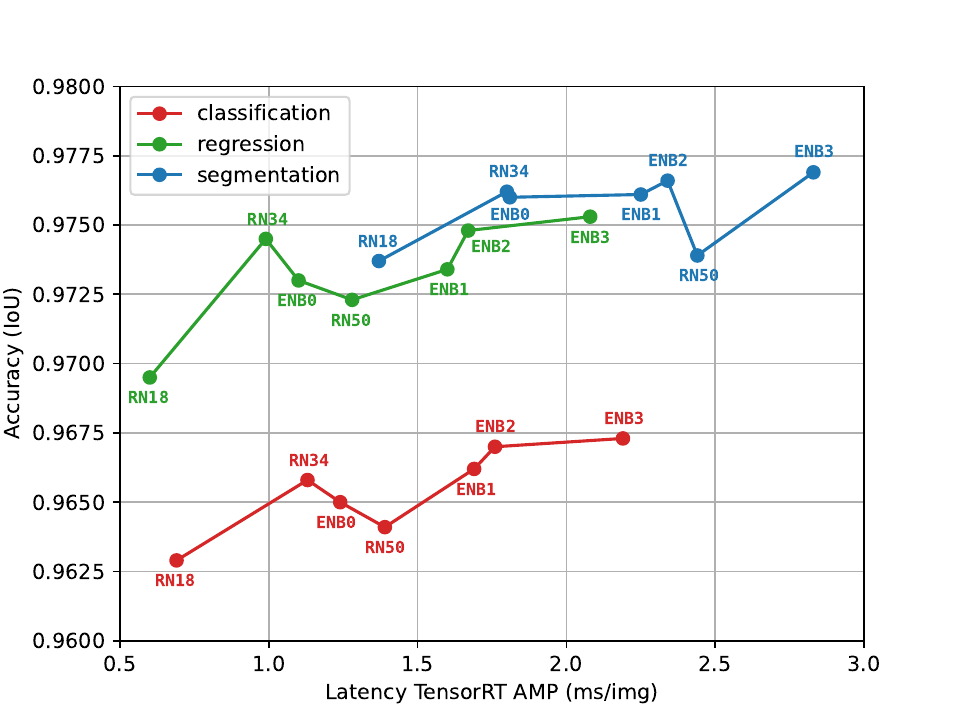}
    \caption{Accuracy versus latency performance across the three paradigms.}
    \label{fig:accuracy_latency}
\end{figure}

Fig. \ref{fig:accuracy_latency} delineates the relationship between accuracy and latency within the classification, regression, and segmentation paradigms, using the same backbone panel as in section \ref{sec:experiments}. From an accuracy perspective, segmentation-based models perform the best, closely followed by regression models, and classification models lay on the lower spectrum. The accuracy range, however, remains notably tight, with a marginal difference of only 1.4 percentage points between the highest performing model (segmentation with EfficientNet-B3) and the lowest (classification with ResNet-18). The superior accuracy observed when employing the segmentation paradigm can be attributed to its intrinsic advantages. U-Net architectures, through decoder upsampling layers, achieve pixel-wise predictions and further benefit from low-level feature maps via skip connections from the encoder. In contrast, classification and regression models rely solely on the final feature maps output by the encoder for their predictions. Furthermore, segmentation models are trained with a region-based loss function (Dice loss) and are evaluated with a similarly region-based metric (IoU), whereas the loss functions for classification and regression models do not align as closely with the evaluation metric.

This accuracy superiority however comes with a cost, as segmentation models are associated with the highest latencies. Regression models are the fastest, followed closely by classification models. Here, the gap is more pronounced, with the fastest regression model being 2.3 times faster than the fastest segmentation model. Overall, a linear trend between accuracy and latency is observed across paradigms; however, models that integrate the ResNet-50 backbone deviate from this trend, delivering lower accuracy than their latency would suggest. This may be attributed to overfitting, a result of the disproportionately large number of parameters relative to the task's complexity and the limited size of the training set.

\begin{table}
    \centering
    \caption{Performance Comparison of the Paradigms}
        \begin{tabular}{lcccc}
        \toprule
        \textbf{Model} & \textbf{Param.} & \textbf{MACs}   & \textbf{Latency PT / TRT} & \textbf{IoU}    \\
        \midrule
        CLS-RN50       & 61.55M          & 21.40G          & 6.12 / 1.39 ms            & 0.9639          \\
        REG-RN50       & \textbf{25.75M} & \textbf{21.36G} & \textbf{6.04 / 1.28 ms}   & 0.9723          \\
        SEG-RN50       & 32.52M          & 42.53G          & 9.86 / 2.44 ms            & \textbf{0.9739} \\
        \midrule
        ResNet-50      & 25.56M          & 21.36G          & 6.12 / 1.28 ms            & -               \\
        \bottomrule
    \end{tabular}
    \label{tab:paradigms_rn50}
\end{table}

Table \ref{tab:paradigms_rn50} details the specifications of the three paradigms when implemented with the ResNet-50 backbone. The regression model is unsurprisingly the lightest and fastest. The classification model is heavily burdened by its last fully connected layer, which outputs a 16512-dimensional vector (2\texttimes64\texttimes129), resulting in a parameter count more than twice that of the regression model with its 129-dimensional output. Yet, this does not translate into an equally large increase in MACs, since these extra parameters are not convolutional; the inference time of the classification model is then only slightly higher than that of its regression counterpart. Conversely, the segmentation model, despite a moderate increase in parameters, sees a doubling in MACs and latency, a consequence of the additional convolutional operations in its decoder layers.

\begin{figure}
    \centering
    \includegraphics[width=0.95\linewidth]{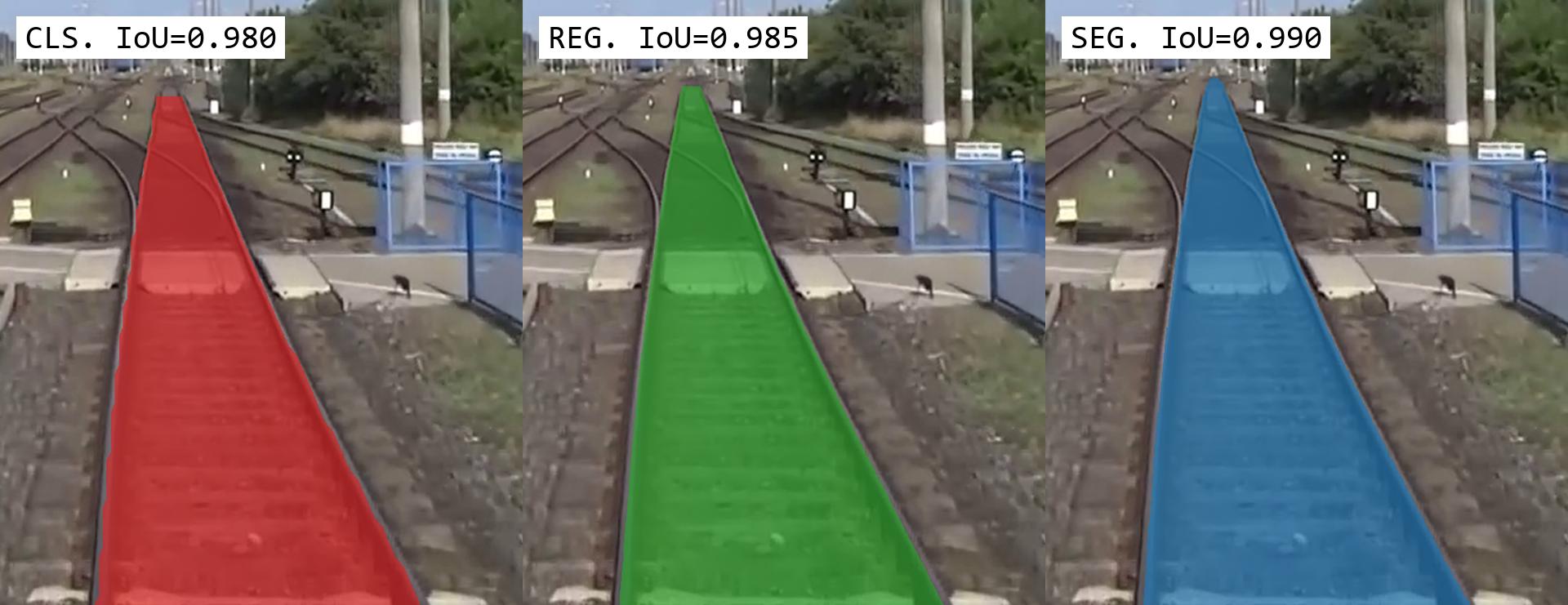}
    \caption{Comparative visualization of ego-path predictions for the three paradigms using the EfficientNet-B3 backbone.}
    \label{fig:qualitative_comparison}
\end{figure}

Fig. \ref{fig:qualitative_comparison} provides a visual comparison of predictions from the three paradigms, using the same input image and equipped with the EfficientNet-B3 backbone. At first glance, the predictions appear identical. Closer inspection of such images reveals that segmentation models' predictions are generally very precise and consistent, whereas the predictions of regression models occasionally deviate by a few pixels from the exact center of the rails. Classification models share this imprecision and are further constrained by the grid resolution, resulting in a staircase effect in their predicted path.

\subsection{Discussion}

\begin{figure}
    \centering
    \includegraphics[width=0.95\linewidth]{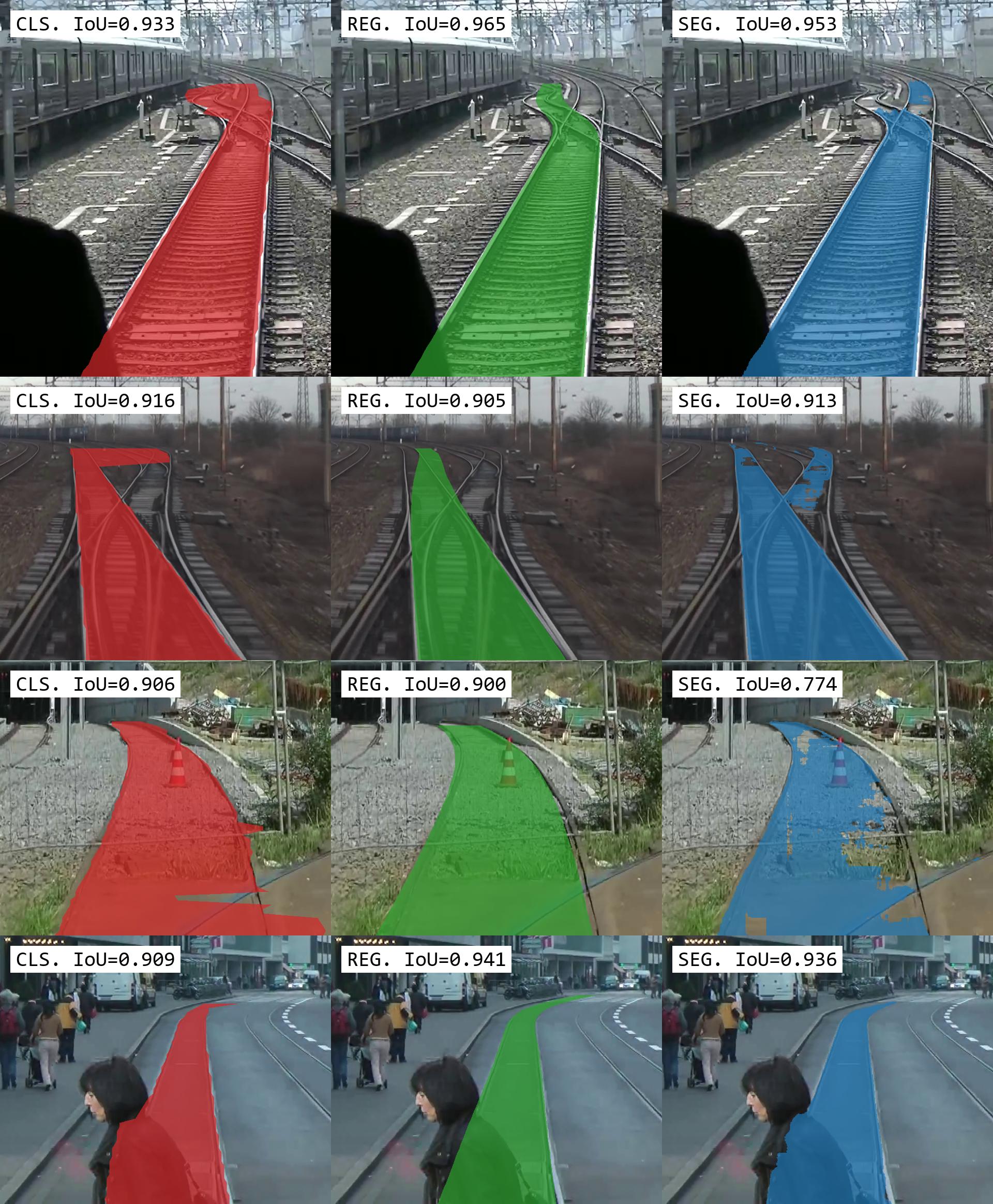}
    \caption{Classification, regression and segmentation models using the ResNet-18 backbone display divergent predictive behaviors when facing challenging scenarios.}
    \label{fig:challenging}
\end{figure}

Segmentation, classification, and regression paradigms embody distinct approaches to ego-path prediction, with nuances that cannot be fully captured by accuracy and latency metrics alone. For a more intuitive grasp of these distinctions, Fig. \ref{fig:challenging} provides a visual comparison of the predictions made by each model in challenging scenarios, using the least accurate backbone (ResNet-18) to accentuate the fundamental differences in their predictive behaviors.

The segmentation approach is the least constrained, offering predictions at the pixel-wise level, whereas the classification and regression approaches conceptualize the ego-path as a set of two continuous rails. This distinction is particularly salient in the upper examples of Fig. \ref{fig:challenging}, where models are hesitant about where to predict the ego-path in the presence of a switch mechanism. The segmentation model is free to predict the path on two different tracks, at the same level of the image, which is structurally impossible for the classification and regression models. While being similarly constrained by their design, these two models exhibit divergent hesitation patterns in their predictive behavior. The classification model's predicted path oscillates between tracks, leading to a somewhat erratic shape. On the other hand, the regression model's hesitation tends to produce a path that lies in between the two tracks, maintaining a consistent curvature amidst uncertainty.

The training objectives of these models elucidate their predictive behaviors. Segmentation models, trained with the Dice loss, and classification models, trained with cross-entropy loss, both optimize functions that process probabilities: for segmentation, the pixel-wise probability of belonging to the path, and for classification, the grid-row level probability distribution. When confident, segmentation models' pixel-wise predictions are close to the extremes of the output value range, whereas under uncertainty, the predictions gravitate toward the middle (threshold) value of the range. Classification models exhibit analogous behavior, outputting probability distributions close to a one-hot vector when confident, and more spread-out probability distributions with several peaks when uncertain between classes. The behavior of these models under uncertainty can lead to significant changes in their predictions, either by transitioning across the threshold value in segmentation models or by shifting from an argmax class to another in classification models. Regression models are instead trained with a loss function that interprets continuous values, and thus, when uncertain, make predictions that lie in the middle ground, averaging the potential outcomes.

The notion of distance in the error between the predicted and the true ego-path is indeed inherently present in the regression paradigm, but absent in the classification and segmentation paradigms. The L1 loss explicitly measures the absolute difference between predicted and actual values, making it sensitive to the magnitude of the error. With the Dice loss, a predicted ego-path region that fails to overlap with the target is penalized, but this penalty does not increase with the distance from the ground truth in the image space. Similarly, the cross-entropy loss does not account for the order or proximity between classes, instead uniformly penalizing the misallocation of probability mass to any wrong class. In the lower examples of Fig. \ref{fig:challenging}, that present unusual scenarios with obstructions, the predictions of the classification and segmentation models appear more volatile than those of the regression model, although the IoU metric does not necessary present significant differences.

Due to its inherent characteristics, the regression approach exhibits a smooth response to uncertainties and perturbations, suggesting a propensity for robustness in fluctuating environments. This attribute is to be considered in the development of auxiliary systems that depend on ego-path prediction, as model resilience can significantly improve the system's overall safety.

\section{Practical Use}

\subsection{Image cropping} \label{sub:img_cropping}

As explained in subsection \ref{sub:data_augmentation}, TEP-Net is trained on cropped images in order to optimize the ego-path information and thus maximize detection precision. For training and testing, the crop coordinates can be derived for each image, based on the ego-path ground truth. However, in real-world applications, such as video inference or real-time operation, the ground truth is not available. Therefore, the crop coordinates must either be predetermined or calculated dynamically.

\begin{figure}
    \centering
    \includegraphics[width=0.9\linewidth]{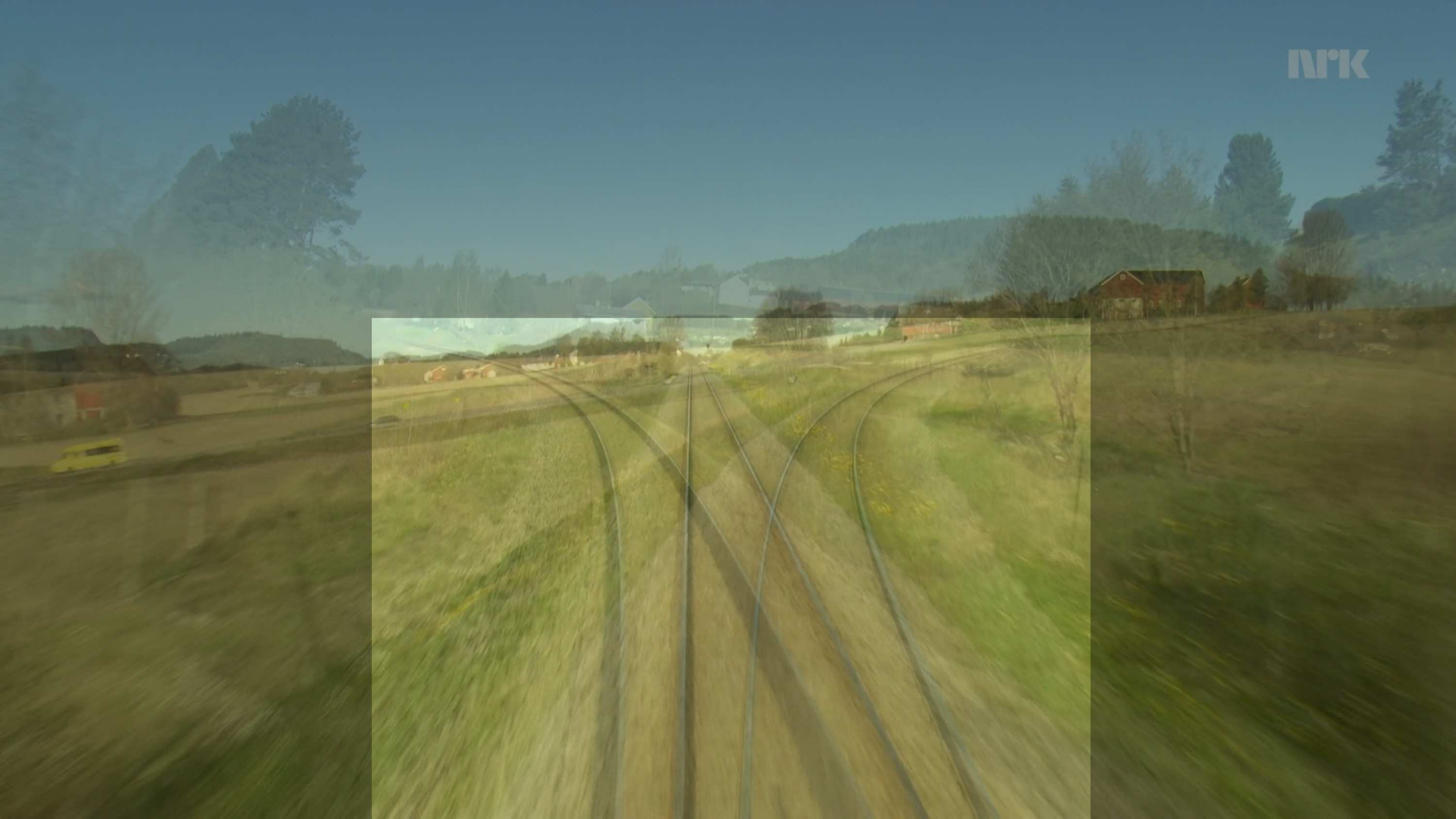}
    \caption{Three superimposed trajectories of a train, highlighting the consistent bounded area of interest for optimized ego-path detection across varying scenarios.}
    \label{fig:bounded_roi}
\end{figure}

An effective crop should encompass the full scope of the object of interest, i.e., the entire ego-path. Fig. \ref{fig:bounded_roi} illustrates three superimposed scenarios (left turn, straight, and right turn) captured from a static camera setup. Despite the divergent ego-paths, the ROI remains within a consistent bounded area. Extensive sections of the input image, notably the sky and peripheral areas (darkened in Fig. \ref{fig:bounded_roi}), do not contribute to ego-path detection, as the ego-path will not be found in these regions. Accordingly, the crop coordinates can be manually fixed for a given video sequence or for a specific train-camera configuration, and applied uniformly across all frames. 

Alternatively, an automatic adaptive cropping approach can be employed: the initial crop encompasses the entire image, and is iteratively refined around the predicted ego-path after each frame processed. Such a dynamic approach allows the crop coordinates to adjust in response to changes in the ego-path, thereby enhancing precision. An implementation for this adaptive cropping strategy is provided in the associated code repository. The algorithm adjusts the crop coordinates based on a running average of the predicted ego-paths. Additionally, it incorporates an anti-collapse mechanism by factoring in the global average of the predictions, which prevents the potential narrowing of the crop area to an impractical extent.

\subsection{Limitations} \label{sub:limitations}

\begin{figure}
    \centering
    \includegraphics[width=0.95\linewidth]{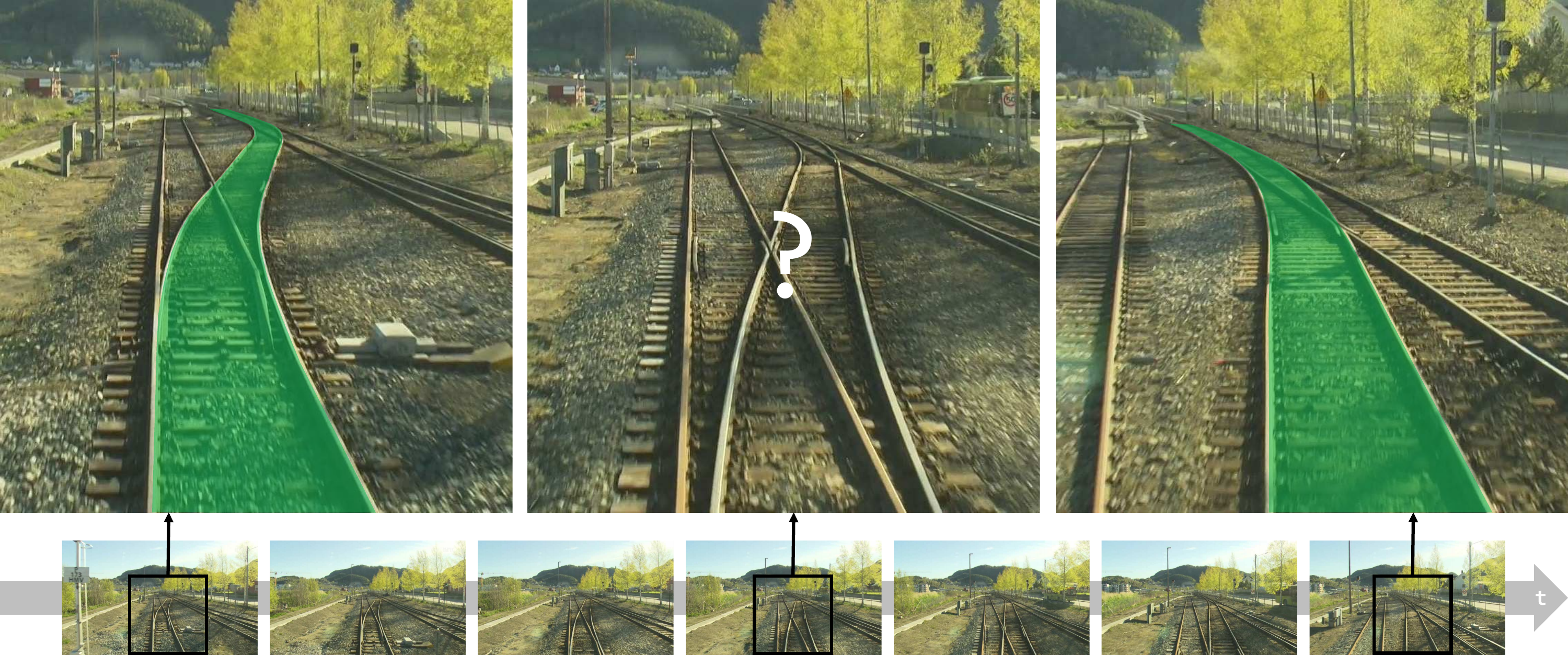}
    \caption{A sequence of frames depicting a train passing over a switch, illustrating a period when the ego-path becomes indiscernible from a single-frame perspective.}
    \label{fig:lost_info}
\end{figure}

The primary limitation of the proposed method stems from TEP-Net being a single-frame-based model, which means it cannot leverage temporal information. This limitation becomes most apparent when the train passes over a switch, as illustrated in Fig. \ref{fig:lost_info}. Initially, the model can accurately predict the ego-path while the switch is fully visible. However, as the train moves forward, the switch gradually exits the frame to the point where it is no longer sufficiently visible for determination, neither by the model nor by a human operator. This initiates a period of uncertainty, which persists until the train has moved far enough from the switch for the path to be clarified through track alignment.

A provisional measure to mitigate this issue could involve monitoring the model's hesitancy, characterized by unexpected fluctuations in the latent and output spaces across consecutive frames, during such scenarios. Additionally, integrating a confidence score into TEP-Net’s predictions would provide a quantifiable indicator of certainty on a single-frame basis. This hesitancy measurement could then serve as an indicator to temporarily suspend reliance on the model until prediction stability is restored. A more comprehensive solution would involve enabling the model to capture temporal context, possibly through the integration of a recurrent neural network component. However, the development of such an enhancement would be facilitated by the open availability of a comprehensive temporal dataset, which is not the case at the time of writing.

Apart from that, enriching the training set with additional complex situations, such as intricate track layouts or occlusions, could perfect the model's ability to handle these scenarios. These might represent the only cases where the model's precision can be significantly enhanced, given that its accuracy in other scenarios is already nearing human-level performance.

\section{Conclusion}

In conclusion, this paper makes a significant contribution to the field of railway intelligent systems by introducing the task of train ego-path detection. This initiative advances the precision and relevance of track detection technologies by focusing on a critical aspect of railway navigation: identifying the immediate path the train is set to follow. Moreover, we successfully tackle this challenge with TEP-Net, a tailored deep learning framework able to accurately and efficiently determine the train's ego-path in a variety of complex railway environments.

Through this work, we also aim to establish a solid foundation for ongoing research and development within the field. By providing open-access to the TEP-Net code and extending the RailSem19 dataset with ego-path annotations, we facilitate further exploration beyond the scope of our study. The comparative analysis of the different paradigms found in track detection literature offers insights that we hope will serve as a valuable resource for future endeavors aimed at advancing railway innovation.

Although incorporating the ability to capture temporal context would undeniably enrich TEP-Net's functionality, the framework as it stands, when paired with an object detection system, already possesses the potential to autonomously identify obstacles threatening a train's path. This capability opens promising avenues toward the development of advanced driver-assistance systems for trains and, potentially, the realization of fully autonomous railway operations.

\section*{Acknowledgment}

This research work contributes to the French collaborative project TASV (Autonomous Passenger Service Train), involving SNCF, Alstom Crespin, Thales, Bosch, and SpirOps. It was conducted in the framework of IRT Railenium, Valenciennes, France, and therefore was granted public funds within the scope of the French program ``Investissements d’Avenir''.

\end{document}